\documentclass[letterpaper, 10 pt, conference]{ieeeconf}  

\IEEEoverridecommandlockouts                              

\usepackage{amsmath} 
\usepackage{graphicx} 
\usepackage{booktabs}
\usepackage{multirow}
\usepackage{amssymb}
\usepackage{pifont}
\usepackage{hyperref}

\usepackage{algorithm}
\usepackage{algpseudocode}

\usepackage{caption}
\title{\LARGE \bf
Compositional Context Fine-Tuning Vision-Language Model for Complex Assembly Action Understanding from Videos
}

\author{Hao Zheng$^{\ast, 1, 4}$, Jinyi Huang$^{4}$, Tiantian Zheng$^{1,2}$, Xun Xu$^{4}$, Tuka Alhanai$^{1,3}$ 
\thanks{$^{\ast}$Hao Zheng is corresponding author: {\tt\small h.zheng@nyu.edu}; $^{1}$Department of Computer Engineering, New York University Abu Dhabi, UAE; $^{2}$Center for Quantum and Topological Systems, New York University Abu Dhabi, UAE; $^{3}$ Center for AI and Robotics, NYUAD, UAE; $^{4}$ Department of Mechanical and Mechatronics Engineering, The University of Auckland, New Zealand. H.Z.: {\tt\small hzhe951@aucklanduni.ac.nz}; J.H: {\tt\small jhua658@aucklanduni.ac.nz}; X.X: {\tt\small x.xu@auckland.ac.nz}. T.A acknowledges support by CAIR and CQTS funded by Tamkeen NYUAD Research Institute Award CG010 and CG008, respectively.}}

\begin{document}

\maketitle
\thispagestyle{empty}
\pagestyle{empty}

\begin{abstract}

Assembly action understanding is a key enabler for effective human-robot collaborative assembly, yet it remains challenging due to subtle motions and fine-grained hand–object interactions. We adapt vision-language models (VLMs) to this challenging domain with Compositional Context Fine-Tuning (CCFT), a method that decomposes assembly actions into semantic elements (\textit{Verb}, \textit{Object}, \textit{Tool}) and fine-tunes VLMs to recognize each action element using templated question-answering pairs. This approach ensures near-deterministic outputs. To enable efficient and effective multi-task learning under limited data, a Layer-Partitioned Alternating Training (LP-AT) method is presented, which assigns distinct model layers to recognize specific action elements through element-specific low-rank adapters. LP-AT alternates weight updates across element-specific adapters, reducing cross-task interference while enabling per-adapter hyperparameter optimization. Furthermore, we create HA-ViD-VQA and IKEA-ASM-VQA datasets from existing assembly video datasets. Extensive experiments on these datasets demonstrate that our method consistently outperforms strong action recognition baselines while providing interpretable element-level predictions that can support diverse downstream applications. Code and dataset are released at \url{https://github.com/x-labs-xyz/CCFT}.

\end{abstract}

\section{Introduction}
\label{introduction}
Building on the strong textual reasoning of large language models (LLMs), vision–language models (VLMs) have advanced multimodal understanding by integrating visual and linguistic cues \cite{10445007,li2025surveystateartlarge}. Multimodal perception enables VLMs to tackle a broader range of real-world challenges, positioning them as promising foundations for human-robot collaboration (HRC), which demands highly precise environmental understanding and physical interaction.

Assembly action understanding represents a critical component of collaborative robotics, as robots must comprehend complex assembly actions to effectively assist humans and acquire manipulation skills through demonstration \cite{10160633,ZHENG2025102976}. However, assembly action understanding poses unique challenges for robotic perception: subtle and intricate actions, distinct actions with similar motions, identical actions with distinct motions, and nuanced hand-object interactions. While VLMs offer considerable potential for advancing complex assembly action understanding, this application domain remains largely unexplored.



Most existing VLM research targets general scene understanding, supporting only basic video summarization or simple question-answering (QA) tasks \cite{nguyen2025}, and lacking specialized adaptions for recognizing complex assembly actions. Moreover, HRC applications demand deterministic outputs, which conflicts with the generative nature of VLM outputs.

This paper proposes two complementary technical ingredients to address these challenges. First, a \emph{Compositional Context Fine-Tuning} (CCFT) method is proposed to decompose assembly actions into semantic elements—\textit{Verb}, \textit{Object}, and \textit{Tool}, and fine-tune VLMs to recognize each element using templated visual question-answering (VQA) pairs. Compared to traditional paradigms, CCFT ensures near-deterministic outputs through templated VQA queries under limited answer spaces and yields interpretable element-level outputs that can support downstream HRC applications. Second, to effectively and efficiently fine-tune VLMs to recognize action elements under limited data, we propose a \emph{Layer-Partitioned Alternating Training} (LP-AT) method. LP-AT assigns disjoint layer groups to individual elements and trains element-specific low-rank adapters (LoRA) \cite {hu2021loralowrankadaptationlarge} applied to specific groups. LP-AT alternately optimizes each element's adapter, thereby confining gradient updates to element-specific parameter subspaces and reducing interference while preserving parameter efficiency. Additionally, LP-AT offers granular hyperparameter tuning (e.g., layer selection, learning rate, rank) flexibility for each adapter.

To validate our approach, we reformulate two existing assembly video datasets, HA-ViD \cite{zheng2023havid} and IKEA-ASM \cite{9423070}, into compositional VQA datasets (HA-ViD-VQA and IKEA-ASM-VQA) with templated QA pairs and action element-level annotations. A state-of-the-art VLM (Qwen2.5-VL \cite{qwen25vl}) is fine-tuned and evaluated on these datasets using our method. The experiments show that our method consistently outperforms strong action recognition baselines while providing interpretable element-level predictions that can support diverse downstream applications.

Our main contributions are:
\begin{itemize}

  \item CCFT is proposed to adapt VLMs for assembly action understanding, which decomposes assembly actions into semantic elements and fine-tunes VLMs to recognize each element, yielding near-deterministic and interpretable element-level predictions.
  \item LP-AT is introduced to facilitate multi-task learning, which assigns disjoint layer groups to subtasks and alternately optimizes task-specific LoRA adapters to reduce cross-task interference and enable per-task hyperparameter tuning.
  \item Two compositional VQA assembly video datasets, HA-ViD-VQA and IKEA-ASM-VQA, are released.
\end{itemize}

\section{Related Work}
\label{related_work}
\paragraph{Assembly Action Understanding for Human-Robot Collaborative Assembly} 
Assembly action understanding is fundamental to enabling effective human-robot collaborative assembly, where robots must comprehend human assembly actions to provide timely assistance, learn manipulation skills, and ensure seamless task coordination \cite{10160633,ZHENG2025102976}. General video action understanding has advanced considerably by leveraging convolutional neural networks \cite{8454294,9219141}, skeleton-based methods \cite{10377200}, and transformer architectures \cite{10377409,10203656}. However, assembly actions present distinct and formidable challenges which include the inherent subtlety and intricacy of manual operations, distinct actions with similar motion patterns, identical actions with distinct motion patterns, and fine-grained hand-object interactions. To address these challenges, researchers have explored enhancing the general action understanding methods by explicitly modeling human-object interactions \cite{10711776} and leveraging detailed human and object pose information \cite{ZHANG2024102659}. However, we note the semantic ambiguity inherent in assembly actions, where visual similarity does not imply semantic equivalence (e.g. tightening a screw or loosening a screw both show a screwdriver turning but semantically they represent opposite actions). This gap motivates our exploration of VLMs in assembly action understanding.

\paragraph{Vision–Language Models for Assembly Action Understanding}
VLMs have extended LLM capabilities to multimodal tasks, enabling joint reasoning across images, videos, and text  \cite{10445007,li2025surveystateartlarge}. Early foundational models, such as CLIP \cite{radford2021learningtransferablevisualmodels} and ALIGN \cite{pmlr-v139-jia21b}, pioneered contrastive image-text alignment, demonstrating strong performance on image-text tasks. More recently, unified multimodal architectures such as LLaVA-Video \cite{zhang2025llavavideovideoinstructiontuning}, and Qwen2.5-VL \cite{qwen25vl} have further extended these capabilities to video understanding \cite{10611454}. However, these models remain primarily designed for general tasks, such as video summarization and simple QA \cite{nguyen2025}, and have been largely underexplored for specialized domains such as assembly video understanding. Given the strong visual perception and reasoning abilities of VLMs, they hold significant potential to address the unique challenges of assembly action understanding. Adapting these powerful generalist models to the manufacturing domain typically requires fine-tuning. Common strategies include full-model updates or parameter-efficient techniques such as LoRA \cite{hu2021loralowrankadaptationlarge} and its variants \cite{dettmers2023qloraefficientfinetuningquantized,liu2024doraweightdecomposedlowrankadaptation}, which are particularly crucial for managing computational costs. Crucially, HRC applications demands exceptionally low tolerance for output ambiguity, challenging VLMs' tendency toward hallucination. Furthermore, while existing assembly video datasets, such IKEA-ASM \cite{9423070}, Assembly101 \cite{9878834}, and HA-ViD \cite{zheng2023havid} lack structured VQA pairs necessary for VLM adaption. 

\paragraph{Compositional Visual Question Answering}
Compositionality, the principle that complex meanings emerge from the combination of simpler elements, is fundamental to human intelligence and essential for computer vision where the scene can be described by their components, including objects, their attributes, and their relationships \cite{10205135}. Building on this principle, researchers have developed compositional VQA approaches for VLMs that explicitly decompose complex visual queries into interpretable subcomponents, employing compositional prompting strategies and causal or scene graph modeling \cite{10655398,parascandolo2025causalgraphicalmodelsvisionlanguage}. Compositional VQA enhances reasoning transparency and supporting generalization to unseen scenes. However, these works remain predominantly focused on image understanding, with limited exploration in video domains. For assembly video understanding, Zheng et al. \cite{10802758} demonstrated the effectiveness of compositional understanding by decomposing actions into four elements (\textit{verb}, \textit{manipulated object}, \textit{target object} and \textit{tool}) and developing separate action segmentation models for each element, subsequently recombining the segmented elements into holistic actions. However, this work relied on traditional approaches and leveraged heuristic reasoning mechanisms to recombine the action elements to form the holistic actions. Given the multimodal reasoning strengths of VLMs, it is promising to apply compositional VQA to assembly action understanding using VLMs.

\section{Methodology}
\label{methodology}
\begin{figure}
\vspace{2mm}
    \centering
    \includegraphics[width=1\linewidth]{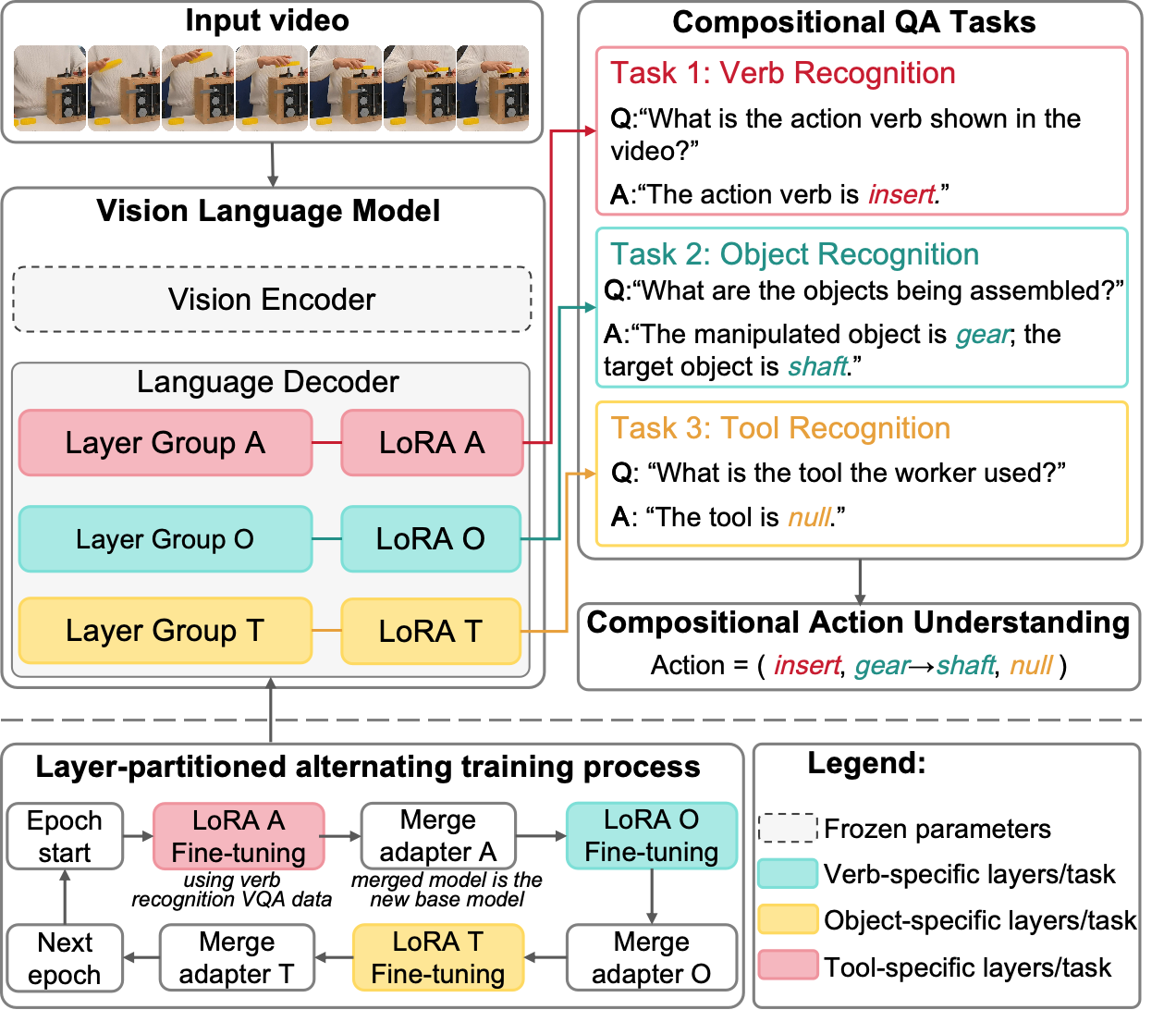}
    \caption{The Overall Framework of Compositional Context Fine-Tuning with Layer-Partitioned Alternating Training.}
    \label{fig:figure1}
\vspace{-3mm}
\end{figure}

Fig. \ref{fig:figure1} presents our overall framework of CCFT and LP-AT. In this section, we formalize the problem (Section \ref{problem_formulation}), describe the construction of compositional VQA datasets (Section \ref{dataset_construction}), introduce the method of CCFT (Section \ref{ccft}) and LP-AT (Section \ref{lpat}), and define the evaluation protocol (Section \ref{evaluation_protocol}).

\subsection{Problem Formulation}
\label{problem_formulation}
We formulate assembly action understanding as a compositional multimodal learning problem. Given an assembly video clip $x = \{f_1, f_2, ..., f_n\}$ where $f_n$ represents the $n$-th frame, our objective is to identify the assembly action $a \in \mathcal{A}$ shown in the video $x$, where  $\mathcal{A}$ is the set of all possible actions. Traditional action recognition methods directly map $x \rightarrow a$, which is a classification problem given the predefined action class set $\mathcal{A}$. In contrast, our method decomposes each assembly action $a$ into semantic action elements: Verb ($v$), Object ($o$), and Tool ($t$). Based on the target application (e.g. on different datasets), the specific action elements can vary. Formally, we aim to train a VLM to generate responses to the queries for each action element, producing $(v, o, t)$, where $(v, o, t)$ can be reconstructed to the corresponding action $a$. Since our approach trains the VLM on VQA tasks, it is not necessary to explicitly provide predefined class sets $\mathcal{V}$, $\mathcal{O}$, and $\mathcal{T}$ for each action element. Instead, the model learns to generate near-deterministic responses in natural language from the limited answer space of each element, providing greater flexibility and scalability.

\subsection{Dataset Construction}
\label{dataset_construction}
We reformulate two existing assembly video datasets, HA-ViD \cite{zheng2023havid} and IKEA ASM \cite{9423070}, into compositional VQA format datasets, HA-ViD-VQA and IKEA-ASM-VQA, facilitating compositional reasoning over fine-grained action elements. Given a video instance $x \in \mathcal{X}$, where $\mathcal{X}$ denotes the entire set of video samples, we define its ground truth action annotation as a structured tuple $a \in \mathcal{A}$, with the composition depending on the dataset's original annotation structure. For HA-ViD, which provides structured action labels with four distinct elements—verb $(v)$, manipulated object $(o_m)$, target object $(o_t)$, and tool $(t)$, we define $a=(v, o_m, o_t, t) \in \mathcal{A}_{HA-ViD}$. In contrast, IKEA ASM contains only verb and object annotations, leading to: $a=(v, o) \in \mathcal{A}_{IKEA}$. Each element $a_i \in a$ is queried independently using a templated natural language question $q_i$, yielding an answer $\hat{a}_i$ from the VLM. This decomposition transforms the holistic action understanding task into a set of compositional QA subtasks, $\mathcal{T}=\{\mathcal{T}_v,\mathcal{T}_o,\mathcal{T}_t\}$, where each subtask $\mathcal{T}_i$ targets the recognition of one action element. Taking an HA-ViD sample for instance:
\begin{itemize}
\item $q_v$: ``What is the action verb shown in the video?'' \\ \quad $a_v$: ``The action verb is \textit{screw}.''
\item $q_o$: ``What are the objects being assembled?'' \\ \quad $a_o$: ``The manipulated object is \textit{nut}. The target object is \textit{bolt}.''
\item $q_t$: ``What is the tool the worker used in the video?'' \\ \quad $a_t$: ``The tool is \textit{wrench}.''
\end{itemize}

\subsection{Compositional Context Fine-Tuning}
\label{ccft}

The VLM model is fine-tuned in a multi-task manner using compositional supervision. Specifically, given a video $x$ and a templated question $q_i$, the model is trained to maximize the conditional likelihood $p(a_i|x, q_i)$, where $a_i$ is the answer. Unlike open-ended VQA, CCFT leverages low-entropy and template-based supervision to minimize ambiguity and improve semantic precision.

Let $\theta_b$ denote the base parameters of the pretrained VLM. To enable multi-tasking and efficient training, a subtask-specific low-rank adapter (LoRA)~\cite{hu2021loralowrankadaptationlarge} $\Delta\theta_i$ is introduced for each subtask $\mathcal{T}_i$, and the adapted model parameters are defined as:
\begin{equation}
\theta = \theta_b + \sum_i \Delta\theta_i
\end{equation}
The training objective for each subtask is to minimize the negative log-likelihood:
\begin{equation}
\mathcal{L}_i = -\mathbb{E}_{(x, q_i, a_i) \sim \mathcal{D}_i} \left[ \log p_{\theta_b + \Delta\theta_i}(a_i \mid x, q_i) \right]
\end{equation}
where $\mathcal{D}_i$ denotes the dataset of annotated QA triplets for subtask $\mathcal{T}_i$. However, naively summing adapters across tasks leads to parameter interference and degrades performance on individual subtasks. To address this, we propose a \emph{Layer-Partitioned Alternating Training} (LP-AT) method, detailed in Section~\ref{lpat}, which isolates task-specific updates by partitioning parameter subsets across model layers during fine-tuning.

Additionally, in this work, Qwen2.5-VL \cite{bai2025qwen25vltechnicalreport} is chosen as the base model. We only fine-tune the language encoder while keeping the vision encoder frozen, because (1) the pretrained vision encoder in Qwen2.5-VL is highly capable; and (2) fine-tuning the vision encoder on our relatively small datasets does not yield significant performance gains.

\subsection{Layer-Partitioned Alternating Training}
\label{lpat}
To enable multi-task adaptation while mitigating cross-task interference, we propose LP-AT. Let \( L = \{l_1, l_2, \ldots, l_N\} \) represent the \( N \) layers of the VLM, partitioned into disjoint task-specific groups \( \mathcal{G} = \{\mathcal{G}_v, \mathcal{G}_o, \mathcal{G}_t\} \), where each \( \mathcal{G}_i \subseteq L \) contains the layers allocated to subtask \( \mathcal{T}_i \), and \( \bigcup_{i} \mathcal{G}_i = L \).

For each subtask \( \mathcal{T}_i \), LP-AT applies a LoRA adapter into the designated layer group \( \mathcal{G}_i \), with rank \( r_i \), learning rate \( \eta_i \), and scaling factor \( \alpha_i \). Let \( \theta_b = \{ W_{\mathcal{G}_i} \}_{\mathcal{G}_i \in \mathcal{G}} \) denote base model parameters, where \( W_{\mathcal{G}_i}  \in \mathbb{R}^{d \times d}\) is a weight matrix at layer group $\mathcal{G}_i$. The adapted weight at each $\mathcal{G}_i$ is defined as:
\begin{equation}
\hat{W}_{\mathcal{\mathcal{G}}_i} = W_{\mathcal{\mathcal{G}}_i} + \alpha_i A_{\mathcal{G}_i} B_{\mathcal{G}_i},  A_{\mathcal{G}_i}\in \mathbb{R}^{d \times r_i},  B_{\mathcal{G}_i} \in \mathbb{R}^{r_i \times d}
\end{equation}

During training, the adapter parameters \( \Delta \theta_i = \{ A_{\mathcal{G}_i}, B_{\mathcal{G}_i} \} \) are optimized by minimizing the subtask-specific loss:
\begin{equation}
\min_{\Delta \theta_i} \ \mathcal{L}_i = -\mathbb{E}_{(x, q_i, a_i) \sim \mathcal{D}_i} \left[ \log p_{\theta_b + \Delta \theta_i}(a_i \mid x, q_i) \right]
\end{equation}

After each update, the adapter is merged into the base model:
\begin{equation}
\theta_b \leftarrow \theta_b + \Delta \theta_i
\end{equation}
and the process proceeds to the next subtask \( \mathcal{T}_{i+1} \). The updated base \( \theta_b \) serves as the initialization for the following step. Once all subtasks are traversed, the next epoch begins.

This alternating method allows each subtask to selectively adapt a distinct subset of the network via ($\mathcal{G}_i$), with specific LoRA rank ($r_i$), scaling factor ($\alpha_i$) and learning rate ($\eta_i$), supporting modular, compositional, and efficient fine-tuning.

\subsection{Evaluation Protocol}
\label{evaluation_protocol}
In manufacturing scenarios, it is essential to ensure unambiguous and deterministic model outputs. Our model outputs the action elements encoded as strings in a structured format (see Section~\ref{dataset_construction}). We adopt a compositional evaluation protocol to assess the discriminative accuracy at both the element and action levels. This protocol facilitates fine-grained analysis of model performance, providing actionable insights for diagnosis and targeted improvement.

Let the ground-truth labels be $y_i$ and the model predictions $\hat{y}_i$ for each action element $i \in \{v, o, t\}$, where $ y_i  $ and  $\hat{y}_i$ are string labels, and they are extracted from the dataset annotation and model output, respectively.

We define the per-element accuracy as:
\begin{equation}
\label{acc_e}
\mathrm{Acc}(i) = \mathbb{I}[\mathrm{sim}(\hat{y}_i, y_i) \geq \tau] \quad \text{for } i \in \{v, o, t\}
\end{equation}
where $\mathrm{sim}(\cdot, \cdot) \in [0, 1]$denotes a normalized string similarity, and \( \tau \) is a similarity threshold.

The predicted holistic action \( \hat{y}_a \) is constructed by combining the predicted action elements \( \hat{y}_i \) (\(i \in \{v, o, t\}\)) following the predefined format. The holistic action accuracy is then defined as:
\begin{equation}
\label{acc_a}
\mathrm{Acc}(a) = \mathbb{I}\left[ \mathrm{sim}(\hat{y}_a, y_a) \geq \tau \right]
\end{equation}
where $y_a$ denotes the ground-truth holistic action string. In our implementation, similarity is computed as the cosine similarity between label embeddings encoded by MiniLM~\cite{wang2020minilmdeepselfattentiondistillation}. \( \tau \) is set to \(0.95 \) to ensure robustness to minor formatting or encoding errors while preserving the near-exact semantic matches.  Although this specific threshold value is an empirical choice that did not impact overall performance metrics in our experiments, the use of a configurable threshold enhances the robustness of the evaluation method.

\section{Experiment}
\label{experiment}
\subsection{Datasets}
\label{datasets}
We select representative subsets from the HA-ViD \cite{zheng2023havid} and IKEA-ASM \cite{9423070} and convert their annotations into our structured VQA format (Section \ref{dataset_construction}), creating HA-ViD-VQA and IKEA-ASM-VQA\footnote{More details in supplementary document: \url{https://github.com/x-labs-xyz/CCFT/tree/main/Supplementary}}.

\textbf{HA-ViD-VQA} captures two-handed assembly actions from three camera viewpoints. The dataset contains 511 left-hand videos (444 train/67 test) and 536 right-hand videos (452 train/84 test) per viewpoint. Each action is annotated with 4 elements: \textit{verb}, \textit{manipulated object}, \textit{target object}, and \textit{tool}. The vocabulary comprises 6 verbs, 38 objects (including manipulated and target objects), and 5 tools, yielding 56 distinct holistic actions. 

\textbf{IKEA-ASM-VQA} focuses on furniture assembly actions captured from a top-view perspective, containing 559 videos with 473/86 train/test splits. Unlike HA-ViD, this dataset provides 2-element annotations (\textit{verb}, \textit{object}) without hand differentiation. The vocabulary includes 12 verbs and 11 objects, forming 24 holistic actions.

\subsection{Implementation Details}
\label{implementation_details}
We use Qwen2.5-VL-7B as the base VLM and fine-tune its language decoder (28 layers). Each video keeps the original frame rate but is temporally downsampled to a maximum of 76 frames, and the processing rate is set to 2 fps, due to GPU memory constraints. For HA-ViD-VQA, we partition the layers 0–9 for verb recognition, 10–23 for manipulated and target object recognition, and 24–27 for tool recognition. This design is motivated by two considerations: (1) the verb $\rightarrow$ object $\rightarrow$ tool progression intuitively follows the compositional order in our action decomposition framework, which also parallels the developmental trajectory in human motor learning: basic motor skills $\rightarrow$ object manipulation $\rightarrow$ tool use \cite{needham_how_2023}, and (2) the layer allocation (10:14:4) corresponds to the estimated relative complexity of each recognition task and has been empirically validated through preliminary experiments (see supplementary document). AdamW optimizers using learning rates of  $5\times10^{-5}$,  $5\times10^{-5}$,  $2\times10^{-5}$ with cosine scheduling are applied to the three adapters respectively. For IKEA-ASM-VQA, layers 0–13 are allocated for verb recognition, and 14–27 for object recognition, with AdamW optimizers using a learning rate of $5\times10^{-5}$ and cosine scheduling for both adapters. All adapters are applied with rank 256 and a scaling factor of 256. Training is conducted for 40 epochs with a per-device batch size of 1, distributed across four NVIDIA RTX A6000 GPUs (48GB each). We use a warmup ratio of 0.1 and gradient clipping at 1.0 to ensure stable training.

\subsection{Main Results}
\label{main_results}
\subsubsection{Comparison with strong baselines}
We benchmark our method against three representative action recognition models: a CNN-based model with temporal shift (TSM \cite{9219141}), a unified transformer-based model (UniFormerV2  \cite{10377409}), and a pre-trained masked autoencoder (VideoMAE V2 \cite{10203656}). All models are evaluated on the HA-ViD-VQA and IKEA-ASM-VQA datasets for the holistic action recognition accuracy. A key methodological distinction is that our approach utilizes VQA annotation files, whereas the baselines employ traditional categorical action labels. Quantitative results comparing the accuracy of our method against the baselines are presented in Table
\ref{tab:table1}. While action recognition studies commonly report both Top-1 and Top-5 accuracies \cite{9219141,10377409,10203656}, our method produces direct semantic outputs rather than ranked scores, making Top-5 evaluation inapplicable. We therefore report only Top-1 accuracy using Equation~\ref{acc_a}.

\begin{table}[h!]
\vspace{2mm}
\centering
\setlength{\tabcolsep}{2pt} 
\caption{Action Recognition Accuracy on HA-ViD-VQA and IKEA-ASM-VQA. The accuracy (\%) for left-hand and right-hand action recognition is reported separately under side, front, and top views for HA-ViD-VQA. Our method outperforms baselines.}
\renewcommand{\arraystretch}{0.9} 
\begin{tabular}{cccc}\toprule
HA-ViD-VQA   & Method      & Left-hand      & Right-hand    \\
             &             & Acc. (\%)        & Acc. (\%)       \\ \hline
             & TSM         & 44.78            & 20.24           \\
Side-view    & UniFormerV2 & 43.28            & 38.10           \\
             & VideoMAEv2  & 41.79            & 36.90           \\
             & Ours        & \textbf{46.27}   & \textbf{47.62}  \\ \hline
             & TSM         & 38.81            & 21.43           \\
Front-view   & UniFormerV2 & 43.28            & 36.90           \\
             & VideoMAEv2  & 40.29            & 29.76           \\
             & Ours        & \textbf{49.25}   & \textbf{42.86}  \\ \hline
             & TSM         & 38.81            & 22.62           \\
Top-view     & UniFormerV2 & 44.78            & \textbf{38.10}  \\
             & VideoMAEv2  & 40.29            & 36.90           \\
             & Ours        & \textbf{53.73}   & \textbf{38.10}  \\ \bottomrule
IKEA-ASM-VQA & Method      & \multicolumn{2}{c}{Accuracy (\%)}  \\\hline
             & TSM         & \multicolumn{2}{c}{25.58}          \\
             & UniFormerV2 & \multicolumn{2}{c}{46.51}          \\
             & VideoMAEv2  & \multicolumn{2}{c}{52.32}          \\
             & Ours        & \multicolumn{2}{c}{\textbf{66.28}} \\ \bottomrule
\end{tabular}
\label{tab:table1}
\vspace{-3mm}
\end{table}


Table \ref{tab:table1} shows that our method consistently outperforms all baselines across both datasets. These results highlight the effectiveness of our compositional fine-tuning strategy for VLMs, which leverages fine-grained VQA annotations to improve assembly action understanding. Notably, right-hand action recognition tends to be less accurate than left-hand due to more complex motion patterns and frequent occlusions caused by right-handed workers \cite{zheng2023havid}. This suggests that future research should explore approaches that better handle visual occlusions to further improve recognition performance.

\subsubsection{Action Element Recognition Performance}
To further evaluate the compositional action understanding capability of our method, we report in Table \ref{tab:table3} the recognition accuracy of individual action elements, as well as the holistic action on HA-ViD-VQA and IKEA-ASM-VQA datasets using Equation \ref{acc_e}.

\begin{table}[h!]
\vspace{-2mm}
\centering
\setlength{\tabcolsep}{2pt} 
\caption{Element-wise Action Recognition Accuracy on HA-ViD-VQA and IKEA-ASM-VQA. For HA-ViD-VQA, we report the recognition accuracy (\%) for each action element: Verb, Manipulated Object (MO), Target Object (TO), Tool, and the holistic Action. LH and RH denote left-hand and right-hand actions, respectively. SV, FV, and TV refer to side, front, and top views. Tool recognition achieves the highest accuracy, while verb recognition consistently outperforms object recognition. For IKEA-ASM-VQA, we report the recognition accuracy (\%) for Verb, Object, and the holistic Action.}
\begin{tabular}{lcccc|c}
\bottomrule
HA-ViD & Verb                      & MO      & TO      & Tool    & Action                    \\
       & $v$                       & $o_m$        & $o_t$   & $t$     & $a$                       \\ \hline
LH-SV  & 61.19                     & 55.22   & 53.73   & 88.06   & 46.27                     \\
LH-FV  & 62.69                     & 58.21   & 53.73   & 95.52   & 49.25                     \\
LH-TV  & 71.64                     & 68.66   & 61.19   & 92.54   & 53.73                     \\
RH-SV  & 64.29                     & 59.52   & 58.33   & 96.43   & 47.62                     \\
RH-FV  & 58.33                     & 52.38   & 51.19   & 96.43   & 42.86                     \\
RH-TV  & 52.38                     & 48.81   & 48.81   & 97.62   & 38.10                     \\ \bottomrule
IKEA   & Verb                      & \multicolumn{3}{c|}{Object} & Action                    \\
       & $v$                       & \multicolumn{3}{c|}{$o$}    & $a$                       \\ \hline
       & \multicolumn{1}{l}{68.60} & \multicolumn{3}{c|}{79.10}  & \multicolumn{1}{l}{66.28} \\ \bottomrule
\end{tabular}
\label{tab:table3}
\vspace{-3mm}
\end{table}

Considering the number of categories for each action element (see Section \ref{datasets}), the results in Table \ref{tab:table3} reveal that recognition accuracy is closely related to task complexity. Tool recognition achieves the highest accuracy, partly because of the small number of tool categories and the fact that only 15 out of 56 holistic actions involve tools. In addition, tools tend to have more visually distinctive appearances compared to the objects in HA-ViD-VQA. On HA-ViD-VQA, verbs consistently outperform object recognition, suggesting that motion patterns provide stronger discriminative signals than (object) appearance for fine-grained assembly actions. Conversely, IKEA-ASM-VQA shows the opposite trend. This reversal likely reflects the different action granularities: HA-ViD captures micro-manipulations requiring precise motion understanding, while IKEA-ASM's furniture assembly involves macro-manipulations on larger, more visually distinct objects that are easier to identify. The gap between element-wise and holistic action accuracy further highlights the challenge of composing multiple element predictions into a coherent action prediction. While decomposition enables interpretable predictions and targeted improvements, the composition mechanism itself needs sophisticated aggregation beyond simple concatenation to handle element uncertainties effectively.

\subsubsection{Qualitative Comparison}
\begin{figure}[!h]
\vspace{2mm}
    \centering
    \includegraphics[width=1\linewidth]{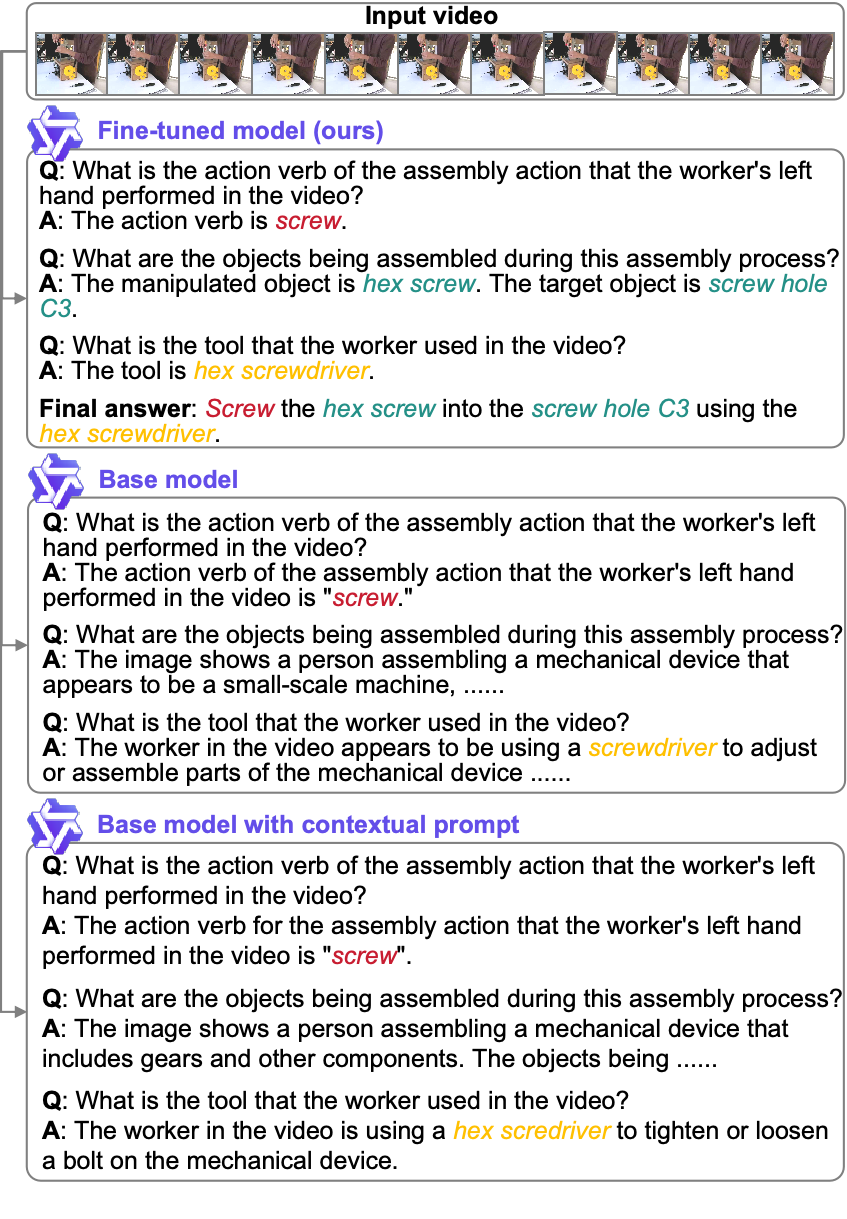}
    \caption{Qualitative Comparison of Assembly Action Understanding Outputs from Our Fine-Tuned Model, the Base Model, and the Base Model with Contextual Prompt. }
    \label{fig:qualitative}
\vspace{-4mm}
\end{figure}

To better understand how CCFT improves assembly action understanding, we compare outputs from three configurations: our fine-tuned model, base Qwen2.5-VL, and base model with contextual prompting (providing lists of possible verbs, manipulated objects, target objects, and tools). Fig. \ref{fig:qualitative} shows results for a video in HA-ViD-VQA with ground truth: \textit{screw the hex screw into the screw hole C3 using the hex screwdriver}. Our fine-tuned model correctly identifies all elements with precise, deterministic outputs suitable for manufacturing applications. In contrast, the base model, despite correctly identifying the verb ``\textit{screw}," generates vague object descriptions (``\textit{mechanical device}," ``\textit{small-scale machine}") and verbose explanations lacking assembly-specific details. Providing contextual prompts yields only marginal improvements—while identifying the tool correctly, the model still produces imprecise object descriptions and verbose outputs. This demonstrates that compositional fine-tuning is essential for learning assembly-specific visual-semantic mappings; simply constraining the output space through prompting is insufficient.

\subsection{Ablation Studies}
\label{ablation_studies}
We conduct ablation studies to analyze the contribution of each key component in our method. All experiments use HA-ViD-VQA due to its comprehensive four-element annotations, using the same configurations in Section \ref{implementation_details}.

\textbf{Compositional vs. Non-compositional Fine-tuning}: We compare CCFT against a baseline that fine-tunes the same VLM on holistic action recognition with no decomposition. Table \ref{tab:table5} compares holistic action recognition accuracy between compositional (CCFT) and non-compositional fine-tuning on HA-ViD-VQA. CCFT delivers consistent gains, with improvements of up to 4.77 percentage points and an average increase of 2.33 percentage points. We attribute these improvements to task decomposition, which reduces label ambiguity and focuses model attention on element-specific visual and semantic cues. Beyond higher accuracy, CCFT generates interpretable intermediate outputs for downstream applications, and its compositional structure permits targeted improvements through element-specific optimization.

\begin{table}[]
\vspace{2mm}
\renewcommand{\arraystretch}{0.9} 
\centering
\setlength{\tabcolsep}{2pt} 
\caption{Comparison of Compositional vs. Non-compositional Fine-tuning for Action Recognition on HA-ViD-VQA. Top-1 accuracy (\%) for left-hand and right-hand action recognition is reported separately under side, front, and top views. Compositional approach improves performance up to 4.77\% and average of 2.33\%.}
\begin{tabular}{cccc}
\toprule
View                   & Method            & Left-hand Acc. (\%) & Right-hand Acc. (\%) \\ \midrule
\multirow{2}{*}{Side}  & Compositional     & 46.27     & 47.62      \\
                       & Non-compositional & 44.78     & 47.62      \\  \hline
\multirow{2}{*}{Front} & Compositional     & 49.25     & 42.86      \\
                       & Non-compositional & 46.27     & 38.10      \\ \hline
\multirow{2}{*}{Top}   & Compositional     & 53.73     & 38.10      \\
                       & Non-compositional & 53.73     & 33.33      \\ \bottomrule
\end{tabular}
\label{tab:table5}
\end{table}

\textbf{Layer Partitioning vs. Shared Adapter}: We compare the layer partition mechanism of LP-AT against a variant that uses a single LoRA adapter shared across all layers and subtasks (see Table \ref{tab:table6}). 
Compared to LP-AT (Table \ref{tab:table3}), the shared adapter consistently yields lower performance, especially in object and holistic action recognition. This degradation can be attributed to the shared adapter’s inability to model the distinct learning dynamics and feature requirements of each subtask, as all subtasks must adapt within a common parameter space. In contrast, our layer-partitioned strategy allows different network layers to specialize for individual subtasks via isolated adapters, improving learning efficiency and reducing representational conflicts. Moreover, since the shared adapter represents a conventional multi-task fine-tuning approach, the observed performance gap suggests that, under limited data, layer-wise specialization may be more effective in learning stable and deterministic representations for fine-grained action understanding.
\begin{table}[h!]
\vspace{-2mm}
    \centering
    \setlength{\tabcolsep}{2pt} 
    \caption{Element-wise Action Recognition Accuracy on HA-ViD-VQA using Shared Adapter. We report the recognition accuracy (\%) for each action element: Verb, Manipulated Object (MO), Target Object (TO), Tool, and the holistic Action. LH and RH denote left-hand and right-hand actions, respectively. SV, FV, and TV refer to side, front, and top views. $\Delta$LA-PT reflects the average gain by applying our method relative to a shared adapter approach, which is superior in object and overall action recognition by 2.68\%-4.77\% and 5.31\% respectively.}
    \begin{tabular}{lcccc|c}\toprule
          &Verb&  MO&  TO&  Tool& Action\\
          &$v$ & $o_m$& $o_t$& $t$& $a$\\
          \midrule
          LH-SV&61.19&  49.25&  47.76&  92.54& 38.81\\
 LH-FV& 65.57& 52.24& 53.73& 95.52&43.28\\
 LH-TV& 73.13& 68.66& 65.67& 92.54&53.73\\
 RH-SV& 67.86& 50.00& 47.62& 92.86&40.48\\
 RH-FV& 61.90& 50.00& 50.00& 97.62&36.90\\
 RH-TV& 67.86& 47.62& 45.24& 95.24&36.90\\ \hline
 $\Delta$LA-PT & -2.30 &	\textbf{4.77} & \textbf{2.68} & -0.42 &	\textbf{5.31} \\
 \bottomrule
    \end{tabular}
    \label{tab:table6}
\vspace{-3mm}
\end{table}

\textbf{Alternating vs. Independent vs. Sequential Training}: We compare three training schemes: (i) \textit{Alternating} (LP-AT): adapters of each subtask trained and merged iteratively within each epoch, enabling cross-task knowledge transfer; (ii) \textit{Independent}: adapters of each subtask trained separately then merged, lacking inter-task learning; (iii) \textit{Sequential}: adapters of each subtask trained one after another with cumulative merging. This comparison isolates the effect of interleaved versus decoupled subtask optimization.

Table \ref{tab:table7} reveals catastrophic performance collapse when merging independently trained adapters. While individual adapters achieve competitive task-specific accuracy before merging, especially in the action elements, the merged model exhibits severe degradation, with an average 29.12\% drop across all elements. This decline arises from parameter interference: each adapter modifies the shared parameter space without awareness of other tasks' requirements, creating conflicting gradients that, when combined, result in a model that fails at all tasks.

Table \ref{tab:table8} illustrates the catastrophic forgetting inherent in sequential training. In this scheme, adapters are trained consecutively (Verb → Manipulated and Target Objects → Tool), with each newly trained adapter merged into the model before proceeding to the next adapter training. We report two critical measurements: performance immediately after training each subtask-specific adapter (showing the adapter's initial performance on the specific subtask) versus performance of the final model after all subsequent training steps. The results reveal severe retroactive interference. While each adapter achieves competitive accuracy when first trained, performance degrades by an average of  46.04\% as subsequent adapters are added. This progressive degradation demonstrates that each new adapter overwrites the representations learned by previous ones, as the model lacks mechanisms to preserve earlier knowledge. 

This ablation study highlights the advantage of our alternating strategy: by iteratively training and merging adapters within each epoch, it learns mutually compatible representations that maintain performance for each subtask.
\begin{table}[h!]
\vspace{-2mm}
    \centering
    \setlength{\tabcolsep}{2pt} 
    \caption{Element-wise accuracy (\%) using independent training scheme on HA-ViD-VQA. Each cell shows performance before/after merging all the adapters.}
    \resizebox{\linewidth}{!}{
    \begin{tabular}{ccccc}\toprule
         &  Verb $v$&  MO $o_m$&  TO $o_t$&  Tool $t$\\\midrule
         LH-SV&  59.70/53.73&  59.70/38.81&  56.72/38.81&  91.04/50.75\\
         LH-FV&  64.18/58.21&  59.70/38.81&  53.73/38.81&  94.03/61.19\\
         LH-TV&  76.12/74.63&  65.67/38.81&  62.69/38.81&  88.06/86.57\\
         RH-SV&  59.52/55.95&  58.33/21.43&  58.33/28.57&  89.29/89.29\\
         RH-FV&  55.95/36.90&  55.95/21.43&  51.19/28.57&  97.62/89.29\\
         RH-TV&  63.10/63.10&  58.33/21.43&  58.33/28.57&  90.48/85.71\\ \hline
 Average drop& 10.23& 49.85& 40.61&15.78\\ \bottomrule
    \end{tabular}
    }
    \label{tab:table7}
\vspace{-3mm}
\end{table}

\begin{table}[h!]
\vspace{-2mm}
    \centering
    \setlength{\tabcolsep}{2pt} 
    \caption{Element-wise accuracy (\%) using sequential training scheme on HA-ViD-VQA. Each cell shows the performance after subtask-specific training/final model performance. The average drop indicates the relative performance decrease of the sequential training scheme.}
    \resizebox{\linewidth}{!}{
    \begin{tabular}{ccccc}\toprule
         &  Verb $v$&  MO $o_m$&  TO $o_t$&  Tool $t$\\\midrule
         LH-SV&  59.70/38.81&  46.27/38.81&  52.24/38.81&  95.52\\
         LH-FV&  64.18/38.81&  58.21/38.81&  56.72/38.81&  95.52\\
         LH-TV&  76.12/38.81&  73.13/38.81&  62.69/38.81&  94.03\\
         RH-SV&  59.52/21.43&  55.95/21.43&  57.14/28.57&  95.24\\
         RH-FV&  55.95/21.43&  55.95/21.43&  50.00/28.57&  97.62\\
         RH-TV&  63.10/21.43&  53.57/21.43&  52.38/ 28.57&  97.62\\ \hline
 Average drop& 52.55& 46.63& 38.95&\\ \bottomrule
    \end{tabular}
    }
    \label{tab:table8}
\vspace{-3mm}
\end{table}

\subsection{Discussion}
\label{discussions}
While our experiments demonstrate the effectiveness of CCFT and LP-AT for assembly action understanding, we identify several limitations and future research directions.
    \textbf{1. Integration into Real-World HRC}: The current work is a foundational step in leveraging VLMs for assembly action understanding. A critical next step is integration into real-world HRC. Future work must focus on optimizing computational efficiency for real-time action recognition, enabling a robot to analyze human actions and provide timely assistance during cooperative assembly tasks.
    
    \textbf{2. Adaptive Hyperparameter Optimization}: Although LP-AT supports subtask-specific hyperparameter configuration (e.g., LoRA rank \( r_i \) , scaling factor \( \alpha_i \) and learning rate \( \eta_i \)), our implementation relied on empirically tuned settings. These hyperparameters may be suboptimal as different action elements exhibit varying learning dynamics throughout training. Future work should explore adaptive hyperparameter optimization strategies that could dynamically adjust hyperparameters for each subtask during training, ensuring stable and efficient learning.
    
    \textbf{3. Principled Layer Partitioning}: On HA-ViD-VQA, tool recognition achieves strong performance with only four layers, whereas verb recognition uses ten layers yet yields lower accuracy. This discrepancy suggests that different subtasks may require distinct network depths and capacities. However, our layer partitioning strategy remains heuristic, while informed by task complexity estimates, lacking theoretical grounding. A promising direction is to conduct layer-wise relevance analysis or develop data-driven layer partitioning methods, informing more principled partitioning decisions.
    
    \textbf{4. Granularity of Action Decomposition}: Our action decomposition stays at the semantic level. Assembly actions inherently involve kinematic constraints and exhibit complex motion patterns, therefore, finer decomposition (e.g. motion-level), deserves attention. Future research should explore multi-level hierarchical decomposition that extends beyond semantic level, enabling hierarchical reasoning and facilitating transfer learning to robotic manipulation tasks \cite{10.1109ICRA}. 
    
    \textbf{5. Dedicated Visual Representation Learning}: Due to the small dataset scale, we froze the visual encoder during training, which is a pragmatic choice that avoids instability. However, as Table \ref{tab:table3} indicates, generic visual features inadequately capture assembly-specific visual patterns or nuanced hand-object interactions. Future efforts should develop assembly-specific vision encoders that emphasize assembly attribute recognition and hand-object interactions.
    
    \textbf{6. Compositional Reasoning}: Our pipeline currently forms the final action predictions by concatenating the predicted action elements. While deterministic, this makes predictions sensitive to single-element errors. Future work should explore robust composition mechanisms, such as lightweight rule-based logic or LLM-driven inference, to reconcile or correct inconsistent element predictions and infer the most plausible holistic action. LLM-based reasoning would also facilitate seamless adaptation of our compositional outputs to new application scenarios \cite{madaan2023self}.


\section{Conclusions}
\label{conclusion}

This paper addressed a critical gap in human-robot collaborative assembly: adapting VLMs for complex assembly action understanding. We presented CCFT with LP-AT, demonstrating that structured task decomposition,  subtask-specific layer allocation, and alternating subtask-specific weight updates are key to successful VLM adaptation in this domain. CCFT stems from two key insights: (1) complex tasks become tractable when decomposed into simpler subtasks; (2) different subtasks demand different network capacities. LP-AT operationalizes these insights by alternately optimizing subtask-specific LoRA adapters, mitigating cross-subtask interference. Beyond quantitative improvements, our method provides interpretable, near-deterministic element-level outputs essential for manufacturing deployment.

To facilitate research in this area, we created the HA-ViD-VQA and IKEA-ASM-VQA datasets as benchmarks for VLM-based compositional assembly action understanding. Extensive experiments and ablations confirm the superiority of our method over state-of-the-art action recognition baselines, validating the benefits of compositional over non-compositional approach, layer partitioning over non-partitioned adapters, and alternating over non-alternating training strategies.





\bibliographystyle{IEEEtran} 
\bibliography{main.bib}

\end{document}